\documentclass[lettersize,journal, twoside]{IEEEtran}

\IEEEoverridecommandlockouts                              

\usepackage[hidelinks,colorlinks]{hyperref}
\hypersetup{
	colorlinks = true, 
	allcolors = blue,
	urlcolor = blue, 
	linkcolor = blue, 
	citecolor = blue, 
}
\usepackage{cite}
\usepackage[amssymb]{SIunits}
\usepackage{graphicx}
\usepackage{subfigure}

\usepackage{subcaption}
\usepackage{multirow}
\usepackage{amsmath,amssymb}

\usepackage{algorithm}
\usepackage{algpseudocode}

\usepackage{enumitem}
\usepackage{balance}
\usepackage[misc]{ifsym}
\usepackage{booktabs}
\usepackage{lscape}
\usepackage{array}
\usepackage{balance}
\usepackage{bm} 
\usepackage{bbm}
\usepackage{algorithm}  
\usepackage{algpseudocode}
\usepackage{multirow}
\usepackage{booktabs}
\usepackage{enumitem}
\usepackage{lscape}
\usepackage[normalem]{ulem}
\usepackage{url}
\usepackage{xcolor}
\usepackage{titlesec}
\setlist[itemize]{noitemsep,nolistsep,leftmargin=17pt}
\setlist[enumerate]{noitemsep,nolistsep,leftmargin=17pt}
\usepackage[font={small}]{caption}
\usepackage[misc]{ifsym}
\newcommand{\fref}[1]{Fig.~\ref{#1}}
\newcommand{\sref}[1]{Section~\ref{#1}}
\newcommand{\tref}[1]{Table~\ref{#1}}

\usepackage[colorinlistoftodos,prependcaption,textsize=tiny]{todonotes}

\title{\LARGE \bf
	iA*: Imperative Learning-based A* Search for Path Planning
}

\titlespacing{\section}
  {0pt} 
  {.5ex plus .2ex minus .2ex} 
  {.5ex plus .2ex} 

\titlespacing{\subsection}
  {0pt}
  {.5ex plus .2ex minus .2ex}
  {.5ex plus .2ex}

\titlespacing{\subsubsection}
  {0pt}
  {.5ex plus .2ex minus .2ex}
  {.5ex plus .2ex}
  
\author{Xiangyu Chen, Fan Yang, and Chen Wang$^{\dagger}$
        \thanks{Manuscript received: May 13, 2025; Revised: July 30, 2025; Accepted: September 27, 2025. This paper was recommended for publication by Editor Aniket Bera upon evaluation of the Associate Editor and Reviewers’ comments. This work was partially supported by the DARPA award HR00112490426 and ONR award N00014-24-1-2003. Any opinions, findings, conclusions, or recommendations expressed in this paper are those of the authors and do not necessarily reflect the views of DARPA or ONR.}
	\thanks{$^{\dagger}$Corresponding Author. The source code and pre-trained models are released at {\tt\small \url{https://github.com/sair-lab/iAstar}}.}
	\thanks{Xiangyu Chen and Chen Wang are with Spatial AI \& Robotics (SAIR) Lab, Institute for Artificial Intelligence and Data Science, Department of Computer Science and Engineering, University at Buffalo, NY 14260, USA. Email: {\tt \{xiangyuc, chenw\}@sairlab.org}}
	\thanks{Fan Yang is with Robotic Systems Lab, ETH Zürich, 8092 Zürich, Switzerland. Email: {\tt fanyang1@ethz.ch}}%
        \thanks{Digital Object Identifier (DOI): see top of this page.}
}

\begin{document}

\maketitle

\markboth{IEEE ROBOTICS AND AUTOMATION LETTERS. PREPRINT VERSION. ACCEPTED
September, 2025}{Chen \MakeLowercase{\textit{et al.}}: iA*: Imperative Learning-based A* Search for Path Planning}

    \begin{abstract}
    Path planning, which aims to find a collision-free path between two locations, is critical for numerous applications ranging from mobile robots to self-driving vehicles.
    Traditional search-based methods like A* search guarantee path optimality but are often computationally expensive when handling large-scale maps.
    While learning-based methods alleviate this issue by incorporating learned constraints into their search procedures, they often face challenges like overfitting and reliance on extensive labeled datasets. 
    To address these limitations, we propose Imperative A* (iA*), a novel self-supervised path planning framework leveraging bilevel optimization (BLO) and imperative learning (IL). The iA* framework integrates a neural network that predicts node costs with a differentiable A* search mechanism, enabling efficient self-supervised training via bilevel optimization. 
    This integration significantly enhances the balance between search efficiency and path optimality while improving generalization to previously unseen maps. 
    Extensive experiments demonstrate that iA* outperforms both classical and supervised learning-based methods, achieving an average reduction of 65.7\% in search area and 54.4\% in runtime, underscoring its effectiveness in robot path planning tasks.
    \end{abstract}
    \section{INTRODUCTION}
    Path planning, determining a route from a starting position to a destination in a given environment, is one of the most basic research topics in robot navigation\cite{review}.
    It is critical for various applications such as those requiring the shortest distance, minimal energy use, or maximum safety~\cite{Quan2020Survey}. 
    
    Due to their optimality, search-based algorithms, such as Dijkstra's algorithm~\cite{dijkstra1959note} and A* \cite{hart1968formal}, have become one of the most popular methods in path planning~\cite{gonzalez2015review}. 
    However, a major drawback of search-based methods is their inefficiency for large-scale maps, since their search space and runtime complexity grow exponentially with the increasing size of maps~\cite{russell2016artificial}.
    Although heuristic strategies such as D*~\cite{dstar} have been introduced to guide more efficient searching directions, their onboard real-time performance is still unsatisfactory, especially on low-power devices and mobile robots. 
	
    \begin{figure}[t]
        \centering
        \includegraphics[width=0.95\linewidth]{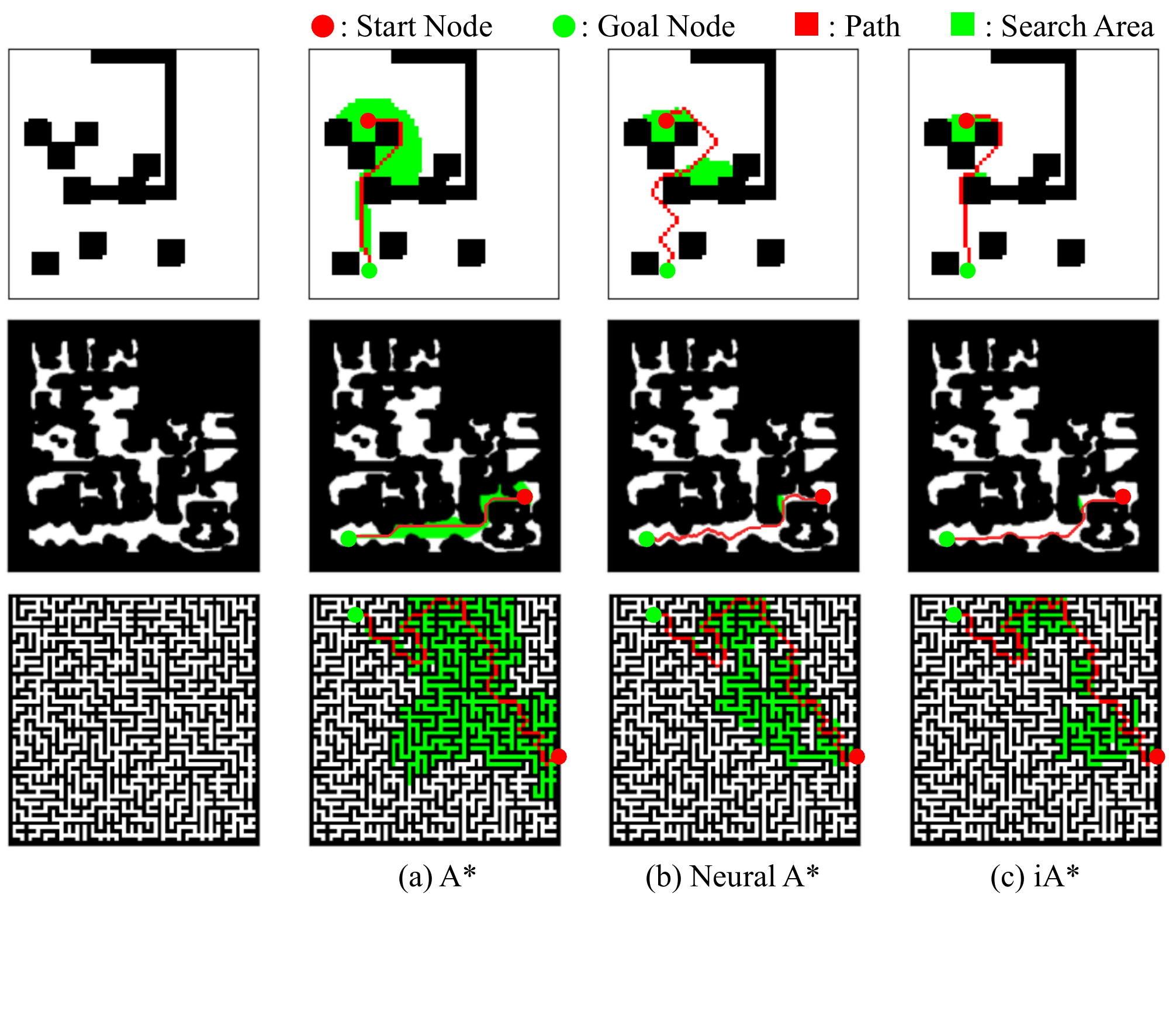}
        \caption{The comparison of the searched area of search-based path planning methods. (a), (b), and (c) display the planning results by using A*, Neural A*, and iA* (ours), where the red and green circles represent the start node and goal node, while the red pixels and green pixels represent the generated path and search area.}
        \label{fig:enter-label}
        \vspace{-10pt}
    \end{figure}

    Recently, researchers have adopted data-driven and supervised learning approaches, such as Neural A*~\cite{yonetani2021nastar} and TransPath~\cite{kirilenko2023transpath}, to predict constrained search space. While these methods lose theoretical optimality guarantees, they have garnered increasing attention due to their ability to significantly reduce the search area and efficiently find near-optimal paths with only a slight increase in path length. However, as they rely on supervised learning, they require extensive pre-processing to get the optimal paths in each training iteration. Moreover, they are prone to overfitting, limiting their effectiveness in out-of-training maps~\cite{gonzalez2015review}.

    To eliminate the requirement for pre-processing and overcome the overfitting challenge, we introduce bilevel optimization (BLO) into search-based path planning, inspired by imperative learning (IL)~\cite{fu2023islam}.
    This results in a novel self-supervised path planning framework, which we refer to as imperative A* (iA*).
    Specifically, iA* consists of an upper-level optimization, which is a neural network to predict a cost for each node, and a lower-level optimization, which is a differentiable A* search engine.
    Different from supervised learning methods like Neural A*~\cite{yonetani2021nastar}, the BLO in iA* enables self-supervised learning by co-optimizing the neural network and the A* search. 
    ~\fref{fig:enter-label} presents a visual comparison between Neural A* and iA*.
    More importantly, iA* enables a more accurate training direction for the neural network by computing a more accurate gradient back-propagating through the bilevel models.
    This leads to a significantly better generalization performance, as demonstrated in both theoretical analysis and practical validation.
    In summary, our contributions are shown as follows:
    \begin{itemize}
        \item We introduce a self-supervised data-driven framework, iA*, for global path planning through bilevel optimization based on imperative learning. iA* enables efficient near-optimal solutions while balancing search area and path length without overfitting via a differentiable self-supervision in the low-level optimization.
        \item We benchmark the performance of various classical and data-driven path planning methods against our proposed iA* framework through extensive experiments on public datasets and simulation environments. The experimental results show that iA* achieves an average of 6.3\% better trade-off between path length and search area, while saving 11.5\% runtime, 3.3\% search areas, and shortening 5.6\% path length than supervised learning methods for unseen maps, demonstrating its significantly better generalization abilities.
    \end{itemize}
	
    \section{Related Works}
    \subsection{Classical Path Planning} 
    The path planning problem, finding a collision-free path from the start node to the goal node with a given map, has been developed for many years. In the beginning, the classical works solve the path planning problem by graph-search approaches~\cite{west2001introduction}, which iteratively search the nodes from the start node to the goal node, such as Breadth First Search (BFS). 
    After that, some algorithms, such as Greedy BFS~\cite{doran1966experiments}, A*~\cite{hart1968formal}, and D*~\cite{stentz1994optimal}, incorporate the heuristic information and provide a search direction to improve the planning efficiency.
    These heuristic search-based algorithms can find the optimal path when their suitable heuristic functions are admissible.
    However, these search-based methods face the problem of low search efficiency with the increasing size of maps due to their exponential time complexity growth.
    
    Unlike search-based path planning methods, sampling-based methods sample nodes and connect nodes to find a solution in the given map. As the number of sampling points increases, the solution is closer to the optimal path. Typically, they stop their sampling when finding a near-optimal path acceptable in the given task to improve their search efficiency. The sampling-based methods can alleviate the computational growth by increasing the dimension of maps. 
    By using a tree structure to store nodes and establish connections between the sampled points, some methods have shown outstanding performance, such as RRT~\cite{lavalle2001rapidly} and its derivatives~\cite{gammell2014informed, gammell2020batch, strub2022adaptively}. Besides, some sampling-based path planning methods~\cite{bohlin2000path, ravankar2020hpprm, liu2023prm} utilize the probabilistic road map to get solutions. However, these sampling-based methods can only get the near-optimal solution and may fail to find paths in complex environments with high obstacle density.
	
    \subsection{Learning-based Path Planning}
    Recently, with the development of deep learning technology, plenty of learning-based path planning methods have appeared. 
    Some of them utilize networks to extract environmental information and provide guidance for improving search efficiency. 
    SAIL~\cite{choudhury2018data} aims to find a suitable heuristic function that could observe and process the whole map instead of the local map during planning. Besides,~\cite{takahashi2019learning} and TransPath~\cite{kirilenko2023transpath} integrate the U-net-based model~\cite{ronneberger2015u} and Transformer-based model~\cite{vaswani2017attention} to learn their heuristic functions, respectively. 
    Some methods~\cite{wang2020neural, yonetani2021nastar} utilize their well-trained network to encode the original map and generate a guidance map. 
    Then, the guidance map allows them to reduce their search area and improve their search efficiency. 
    MPNet\cite{qureshi2020motion} uses two networks, Encoder net (Enet) and Planning net (Pnet), for the path planning task, where Enet handles the environment information, and Pnet can provide the next state of the robot under the current situation.
    
    Some researchers use reinforcement learning (RL)~\cite{kaelbling1996reinforcement} to solve the path planning task. 
    They aim to find an end-to-end solution for the path planning task without any policies.
    They formulate the path planning problem as a Markov decision process where the next state in the path depends on the current state and surroundings. 
    \cite{panov2018grid} demonstrates that it is possible to solve the path planning problem in grid maps using the RL-based method. 
    \cite{gao2019global} combines the path graph with Q-learning, efficiently finding the optimal path.
    However, these learning-based methods face problems of lacking interpretability and extensive data labeling.
	
    \subsection{Imperative Learning}
    Imperative learning (IL)~\cite{wang2025imperative} is an emerging self-supervised neuro-symbolic framework based on bilevel optimization, where the upper-level optimization is a data-driven model and the lower-level optimization is an interpretable reasoning engine.
    IL adopts a self-supervised manner.
    The results from lower-level optimization can provide an adaptive and precise gradient for the training of the data-driven model within upper-level optimization.
    For example, iPlanner~\cite{yang2023iplanner} utilizes a B-spline interpolation algorithm to guide the network training, generating safe and smooth trajectories with depth images. iSLAM~\cite{fu2023islam} adopts the IL framework to integrate the front-end and back-end of visual odometry. The pose estimation guides the training to generate the poses aligned with geometrical reality.
    iMatching~\cite{zhan2024imatching} utilizes the bundle adjustment module to train a neural model, generating the optimal translation matrix for pose estimation. 
    iKap~\cite{li2025ikap} leverages a differentiable model predictive control to achieve a kinematics-aware vision-to-planning framework. These works demonstrate the efficiency of IL in robot navigation tasks. 
    To the best of our knowledge, iA* is the first method using the IL to solve global path planning problems.
	
    \section{Methodology}
    Traditional search-based methods can guarantee finding the optimal paths, while deep neural networks have the ability to reduce the search area by extracting the map information and making predictions. To leverage the advantages of both learning-based and search-based path planning methods, the proposed imperative A* (iA*) adopts the imperative-learning (IL) strategy, including learning-based upper-level optimization and A* based lower-level optimization. 
    
    \subsection{Framework} 
\begin{figure}[t]
        \centering
        \vspace{2pt}
        \includegraphics[width=0.95\linewidth]{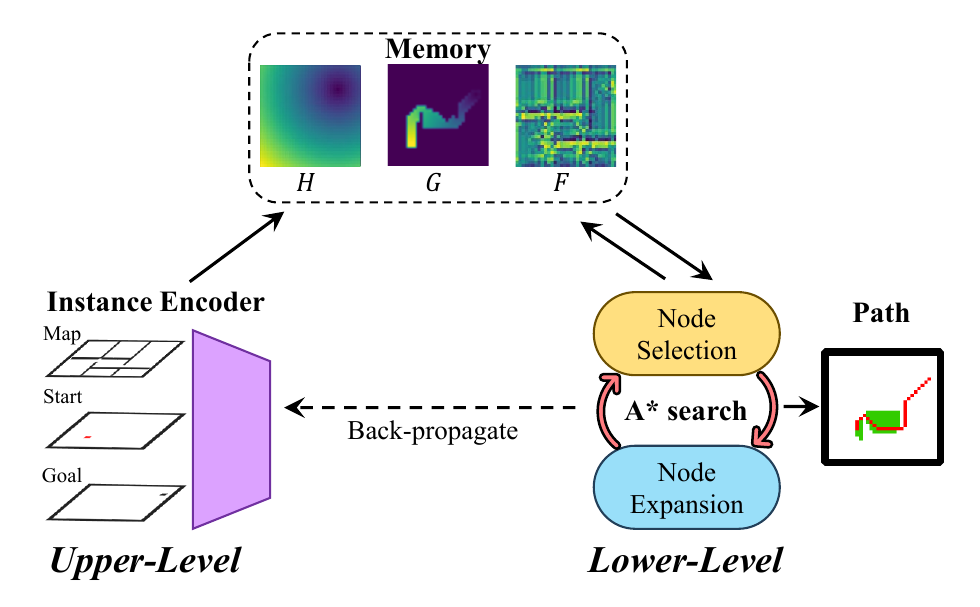}
        \caption{The framework of iA* consists of three main components: (1) an upper-level optimization module featuring an instance encoder that processes the map, start, and goal, (2) a lower-level optimization module that integrates a differentiable A* search for node selection and expansion, and (3) a memory storage module that maintains intermediate representations, $H$, $G$, and $F$ to facilitate interaction between the upper and lower level optimizations.}
        \vspace{-10pt}
        \label{fig:framework}
    \end{figure}

    The framework of iA* is depicted in \fref{fig:framework}, which takes a planning instance as input, a three-layer tensor containing the information of obstacles, start, and goal positions, and outputs the near-optimal path (red) and search area (green).
    The iA* framework can be formulated as a bilevel optimization:
    \begin{subequations}\label{eq:iastar}
        \setlength{\belowdisplayskip}{2pt}
        \begin{align}
            \min_{\theta} \quad & U\left(f(\theta, \bm{x}), \mu^*)\right), \label{eq:high-iastar} \\
            \textrm{s.t.} \quad & \mu^* = \arg\min_{\mu} L(f(\theta), \mu, M), \label{eq:low-iastar}
        \end{align}
    \end{subequations}
    where $\bm{x}$ denotes the input instance, including the map, start node, and goal node, $\theta$ represents the network parameters, and $f(\theta, \bm{x})$ is an instance encoder, and $\mu$ is a set of paths in the solution space.
    Specifically, Eq. \eqref{eq:low-iastar}, defined in \sref{sec:lower-level}, is a lower-level optimization to find the optimal path given a search area predicted by the instance encoder $f$, while Eq. \eqref{eq:high-iastar}, defined in \sref{sec:upper-level}, represents the upper-level optimization for the instance encoder, aiming to find the optimal balance between search area and path length. The memory module contains the intermediate information and establishes a bridge between the two optimizations. 
    
    The main difference between previous methods, like Neural A* \cite{yonetani2021nastar}, and iA* is that the bilevel optimization \eqref{eq:iastar} incorporates a metric-based optimization to provide a more accurate training direction for the network $f$, leading to a more ``generalizable'' smaller search area.
    Intuitively, if the lower-level optimization \eqref{eq:low-iastar} is removed and $\mu^*$ represents constant optimal paths (labels), then bilevel optimization \eqref{eq:iastar} degrades to a single-level optimization and iA* becomes Neural A*, and then the network gradients $\nabla_{\theta} \mathcal{U}$ are
    \begin{equation}
        \nabla_{\theta} \mathcal{U} = \frac{\partial \mathcal{U}}{\partial f_\theta} \frac{\partial f_\theta}{\partial \theta} \quad \text{(Constant Labels $\mu^*$)}.
        \label{eq:gradient_pro1}
    \end{equation}
    In this case, these constant optimal paths restrict gradient propagation and increase the risk of overfitting because such supervised training process only enforces the predicted search area similar to the optimal paths instead of ``facilitating the searching by providing suitable updating guidance''. As a comparison, the gradient of \eqref{eq:iastar} with the chain rule is
    
    \begin{equation}
        \nabla_{\theta} U = \frac{\partial U}{\partial f_\theta} \frac{\partial f_\theta}{\partial \theta} + \boxed{\frac{\partial U}{\partial \mu^*}\frac{\partial \mu^*}{\partial L} ({\frac{\partial L}{\partial f}\frac{\partial f} {\partial \theta}} + \frac{\partial L}{\partial \mu} + \frac{\partial L}{\partial M})},
        \label{eq:gradient_pro2}
    \end{equation}
    
    where the $\boxed{\text{second term}}$ illustrates an important difference.
    That means, the training of iA*, in the $t$-th iteration, relies on the optimal path $\mu_t$ for a search area given by $f(\theta_t)$.
    Since the searching process defined by \eqref{eq:low-iastar} contributes to the gradient calculation, the network tends to predict a lower searching-cost area (lower-level optimization cost $L$) and thus more ``likely'' converge to better solutions. Nevertheless, this process requires a differentiable lower-level optimization, thus we next illustrate its details.

    \subsection{Lower-level Optimization}\label{sec:lower-level}
    The lower-level optimization module is designed to achieve two objectives: (1) executing the searching process for path planning, and (2) providing differentiable computational gradients for back-propagation in upper-level network training.
    The classical A* algorithm inherently relies on a discrete graph structure, presenting challenges in effectively providing differentiable losses for upper-level optimization.
    To overcome this critical challenge, the lower-level optimization employs differentiable A* (dA*)~\cite{yonetani2021nastar} as its search engine, back-propagating the loss of the path planning process.

    In dA*, all variables and parameters are represented in matrix formulations. 
    Specifically, denote the open-list matrix $O$ and closed-list matrix $C$, which are binary and of the same size as the given map, to record the candidate nodes under evaluation and the nodes that have already been evaluated, respectively.  
    The start node $n_s$, the goal node $n_g$, and the currently selected node $n'$ are separately represented by one-hot indicator matrices $N_s$, $N_g$, and $N'$, where only a single element is set to $1$, and all others are $0$.
    Denote $s(n)$ as the accumulated path length from the start node and $h(n)$ as the estimated heuristic distance to the goal.
    The matrices $S$ and $H$ store the $s$ and $h$ values of each node.
    During the path planning process, dA* iteratively selects node $n'$ with the minimum $s(n) + h(n)$ from the open list $\mathcal{O}$ and updates the $S$ matrix accordingly, until the optimal path is found. 
    
    In Neural A*, the neural network estimates the initial $s$ values for each node to accelerate the searching process.
    However, this manner results in error propagation: inaccuracies in $s(n)$ estimation during earlier iterations accumulate and affect subsequent search as the $s$ values are updated relying on the prior $s$.
    To eliminate this problem, we introduce a prediction factor $p$ in iA* to independently represent the influence of upper-level optimization in the Node Selection part of dA*. 
    {$P$ is the matrix storing the predicted $p$ values of each node.}
    Then, we iteratively select the node with minimum $s(n) + h(n) + p(n)$ in the $\mathcal{O}$.
    It is worth noting that if the $p(n)$ equals $(\omega - 1)h(n)$, our A* search algorithm equals weighted A* search~\cite{pohl1970heuristic}, where the $\omega$ is a pre-set weight factor amplifying the influence of the heuristic function.
    In iA*, the node selection is formulated as:
    \begin{equation}
        N' = \mathcal{L}_{\max}{\left(\frac{\exp(-(S+H+P)/\tau)\odot O}{\langle \exp(-(S+H+P)/\tau), O\rangle}\right)},
    \end{equation}
    where $A\odot B$ is the Hadamard product~\cite{horn2012matrix}, $\tau$ is an empirically defined parameter, the same as dA*, and $\mathcal{L}_{max}(A)$ is the function to obtain the $argmax$ index matrix of $A$.
	
    \subsection{Upper-level Optimization}\label{sec:upper-level}
    The upper-level optimization aims to increase the search efficiency and keep the sub-optimality. 
    This objective also balances the search area and path length.
    We take advantage of a neural network in the upper-level optimization because of its great power in information extraction and prediction.
    The neural model is a widely-used U-Net~\cite{ronneberger2015u}, where the decoder is used for feature extraction and the encoder for prediction.
    The use of U-Net ensures that the shape of the predicted $P$ is consistent with the shape of the input maps, avoiding changing the structure of the network and retraining in new environments with different shapes, and providing convenience in robot applications.

    \begin{figure}
        \centering
        \includegraphics[width=0.95\linewidth]{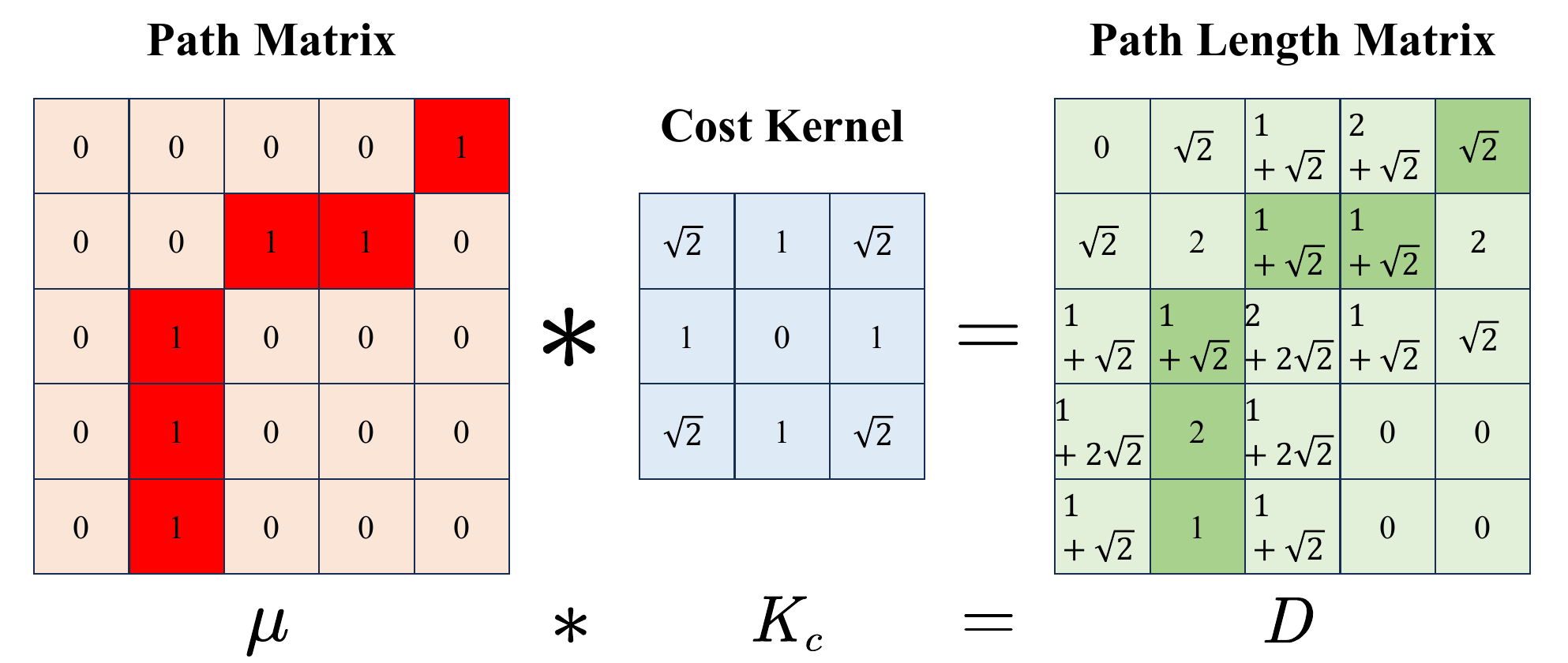}
        \caption{The calculation of the path length. $\mu$ is the Path Matrix generated from the lower-level optimization, $A * K$ represents the convolution operation between matrix $A$ and kernel $K$, and $D$ is the distance matrix for further path length calculation.}
        \label{fig:pathlength}
        \vspace{-10pt}

    \end{figure}
    
    Besides, to avoid the problem of data labeling and overfitting, the training process adopts a self-supervised manner.
    The loss function combines the extracted path length and search area directly from the results of lower-level optimization, which can be formulated as follows:
    \begin{equation}
        U(\mu) =  w_a \mathcal{L}_a(\mu) + w_l \mathcal{L}_l(\mu),
    \end{equation}
    where $U(\mu)$ is the target function of upper-level optimization given $\mu$, $\mathcal{L}_a(\mu)$, $\mathcal{L}_l(\mu)$ are the search area loss and path length loss of path $\mu$, and $w_a$ and $w_l$ are the weights to adjust the ratio of the losses.
    We define $\mathcal{L}_a(\mu)$ as the number of extra visited nodes, which are visited by the algorithm but are not part of the final planned path, in one single path planning process. The calculation is shown as follows:
    \begin{equation}
    \mathcal{L}_a(\mu) = \sum (C_{\mu} - \mu),
    \end{equation}
    where $C_{\mu}$ is the closed-list matrix of lower-level optimization, recording all visited nodes when finding the path $\mu$.

    To accelerate the calculation of path length $\mathcal{L}_l(\mu)$, we employ the widely-used convolution operator, as shown in \fref{fig:pathlength}.
    Specifically, with a constant kernel $K_c = [[\sqrt{2},1,\sqrt{2}]^T, [\sqrt{2},0,\sqrt{2}]^T, [\sqrt{2},1,\sqrt{2}]^T]$ and the path matrix $\mu$, we are able to obtain the distance matrix $D$ as:
    \begin{equation}
        D(i,j)=\mu*K_c=\sum_0^2\sum_0^2K_c(m,n)\cdot\mu(i+m,j+n),
     \end{equation}
    where $D(i,j)$ is the sum of the distances between the node $(i,j)$ and its neighboring path nodes. The path nodes are marked as red squares within Path Matrix in \fref{fig:pathlength}.
    To get the path length $\mathcal{L}_l(\mu)$, the distance values in $D$ are counted according to the path $\mu$, which can be formulated as follows:
    \begin{equation}
        \mathcal{L}_l(\mu) = \frac{\langle D, \mu \rangle}{2},
        \label{eq:length}
    \end{equation}
    Finally, introducing the optimal path $\mu^*$ generated from lower-level optimization, the target function of the upper-level optimization, namely the loss function in network training, is mathematically represented as:
    \begin{equation}
        U(\mu^*) =  w_a \mathcal{L}_a(\mu^*) + w_l \mathcal{L}_l(\mu^*),
    \end{equation}
    where the $w_a$ and $w_l$ are set to 1 empirically for keeping the optimal path while reducing the search area.
    
    In iA*,  the upper-level optimization updates parameters in the embedded network to predict the $P$, pursuing the balance between search area and path length while maintaining near-optimal path quality.
    Unlike the other data-driven methods~\cite{yonetani2021nastar, kirilenko2023transpath}, directly calculating loss with constant optimal paths, our method dynamically calculates loss from the path planning results of lower-level optimization in each iteration.
    Thus, the network parameters are updated to find the optimal balance between search area and path length under temporary conditions, instead of invariably following the constant optimal paths.
    To validate this bilevel training process, we compare the effects of the conventional optimal-path-based training process with ours in the ablation study.

    \begin{figure*}[t]
        \centering
        \vspace{3pt}
        \includegraphics[width=0.99\linewidth]{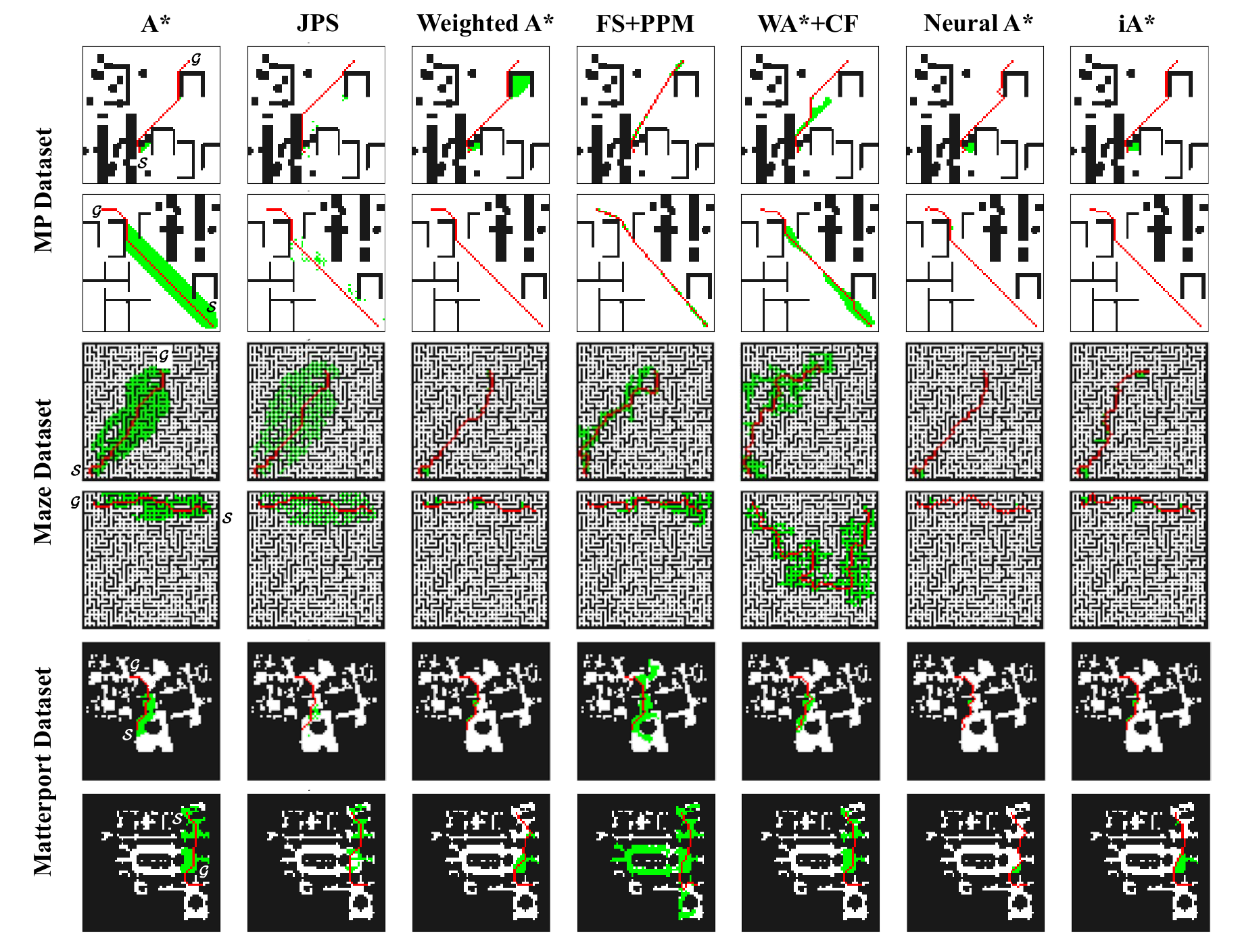}
        \caption{Selected planning results of different search-based path planning methods in the three datasets, MP dataset (the first two rows), Maze dataset (the middle two rows), and Matterport3D dataset (the last two rows). $\boldsymbol{\mathcal{S},\mathcal{G}}$ indicate the start and goal positions in the given map, respectively. The red pixels indicate the generated path and the green pixels indicate the searched area. }
        \vspace{-25pt}
        \label{fig:exp:benchmarks}
    \end{figure*}
    
    \section{Experiments}
    \label{sec:experiments}
    We conduct experiments to evaluate the performance of iA* on path planning problems across diverse environments.
    To ensure a comprehensive comparison, we select the latest data-driven search-based methods as our benchmarks and widely used datasets to provide experimental instances across diverse environments.
    Furthermore, to assess the performance of the iA* algorithm in robot navigation tasks, we conduct simulation experiments in simulated scenarios with a mobile robot.
    The ablation study focuses on the impact of IL strategy and network choice on network training. 

    \subsection{Datasets and Benchmarks}
    To provide training data and diverse test instances for evaluation, we select three publicly available datasets representing distinct environmental configurations. \textbf{Motion Planning Dataset (MP Dataset)}~\cite{bhardwaj2017learning}, \textbf{Maze Dataset}~\cite{maze-dataset}, and \textbf{Matterport3D Dataset}~\cite{chang2017matterport3d}.
    These datasets are widely used for evaluating path planning methods.
    The details about the selected datasets are shown as follows: 
    \begin{itemize}
        \item \textbf{Motion Planning Dataset (MP Dataset)} \cite{bhardwaj2017learning}: This dataset contains a set of binary maps about eight diverse environments with simple obstacle settings.
        \item \textbf{Maze Dataset} \cite{maze-dataset}: This dataset aims to facilitate maze-solving tasks and contains several types of mazes more sophisticated than maps in the MP dataset.
        \item \textbf{Matterport3D Dataset} \cite{chang2017matterport3d}: This dataset provides several Truncated Signed Distance Function (TSDF) maps of indoor scenarios. To get the obstacle setting of these maps, we set a threshold as 0.2 $m$ to select the obstacles. 
    \end{itemize}
    We use the instance generation mentioned in \cite{yonetani2021nastar}, randomly sampling the start and goal positions in the given maps and generating the instances for training and testing. 
    
    To evaluate the path planning performance of the proposed iA* framework, we selected the latest open-source data-driven methods, Neural A* ~\cite{yonetani2021nastar}, Focal Search with Path Probability Map (FS + PPM)~\cite{kirilenko2023transpath}, and Weighted A* + correction factor (WA* + CF)~\cite{kirilenko2023transpath}, as competitors. 
    Besides, we also select the Jump Point Search (JPS)~\cite{harabor2011jps} and Weighted A* (W*)~\cite{pohl1970heuristic} as classical competitors. The JPS algorithm improves the search efficiency by skipping over large sections of nodes through directional pruning and identifying critical jump points, and Weighted A* biases the search toward the goal area to improve search efficiency.
    The Neural A*, FS + PPM, and WA* + CF have been trained by the MP dataset in a supervised fashion, given the optimal solution as labels, and we use the well-trained parameters published on their website.
    For fair comparisons, all the selected data-driven methods are trained exclusively on the MP dataset. The Maze and Matterport3D datasets are used solely for evaluation and remain unseen during training. 
    For the training settings, like learning rate and optimizer, we follow the settings in \cite{yonetani2021nastar, kirilenko2023transpath} and empirically adjust them by extensive training trials.
    
    \begin{table*}[t]
        \centering
        \caption{Experimental Results in the Selected Datasets}
        \label{tab:exp:benchmarks}
        \renewcommand{\tabcolsep}{0.25pt}
        \resizebox{\textwidth}{!}{
        \begin{tabular}
        {p{15pt}>{\centering}p{35pt}>{\centering}p{40pt}>{\centering}p{40pt}>{\centering}p{42pt}>{\centering}p{42pt}>{\centering}p{40pt}>{\centering}p{40pt}>{\centering}p{42pt}>{\centering}p{42pt}>{\centering}p{40pt}>{\centering}p{40pt}>{\centering}p{42pt}>{\centering\arraybackslash}p{42pt}}
        \toprule
        \multirow{2}{*}{\textbf{Size}}&\multirow{2}{*}{\textbf{\;Method}}&\multicolumn{4}{c}{\textbf{MP}} &\multicolumn{4}{c}{\textbf{Maze}}&\multicolumn{4}{c}{\textbf{Matterport3D}} \\
        \cmidrule{3-14}
        &&\textit{Exp}$\uparrow$&\textit{Rt}$\uparrow$&\textit{AL}$\downarrow$&\textit{PL}$\downarrow$
        &\textit{Exp}$\uparrow$&\textit{Rt}$\uparrow$&\textit{AL}$\downarrow$&\textit{PL}$\downarrow$
        &\textit{Exp}$\uparrow$&\textit{Rt}$\uparrow$&\textit{AL}$\downarrow$&\textit{PL}$\downarrow$\\
        \midrule
        \multirow{6}{*}{\textbf{64}} &A*  &0(0) &0(0) & 85.9(17.8)& 73.0(3.1)&
        0(0) &0(0) &142.9(31.9)&131.0(31.3)&
        0(0) &0(0) &42.4(5.2)& 37.2(4.6)\\
        &JPS &\textbf{95.1}(5.9) &- &\textbf{77.2}(22.3) &73.0(3.1)
        &-103(26.7) &- &147.9(107.9)&131.0(31.3)
        &-15.2(19.2) &- &\uline{42.6}(18.7)&37.2(4.6)\\
        &WA* & 64.8(3.0)& 47.7(6.5)& 82.6(20.7)& \uline{75.0}(3.7)
        &43.9(10.4)& 33.2(14.0)& \uline{144.6}(32.2)&\uline{135.7}(31.1)
        &48.1(13.8)& 36.1(31.3)& 42.9(5.2)&39.1(4.7)\\
        &Neural A*&63.5(2.6) &39.6(5.4) &86.8(20.4) & 79.7(3.7)
        &40.6(13.1) &\uline{37.5}(19.2) &151.0(30.9)&142.7(29.7) 
        &42.3(13.4) &10.0(60.2) &44.4(5.7)&40.8(5.4)\\
        &WA*+CF &68.8(2.6) &48.0(7.2) &85.6(22.3)&78.7(4.1)
        &\textbf{56.8}(11.1) &\textbf{42.7}(16.8) &149.2(30.0) &141.5(29.0)
        &\uline{57.2}(16.0) &\textbf{45.6}(22.3) &{45.6}(6.2)&42.1(6.0)\\
        &FS+PPM&\uline{89.0}(0.7) &\textbf{88.8}(0.9) &\uline{78.0}(16.0) &\textbf{73.8}(3.1)
        &-7.9(10.9) &-15.0(31.0)&149.6(30.4)&139.1(29.6)
        &9.3(26.1)&-17.4(56.3)&42.7(5.3)&\uline{38.3}(5.1)\\
        &iA*(Ours) &{69.4}(2.9) &\uline{50.4}(5.7)&{82.3}(19.0)& \uline{75.0}(3.4)
        &\uline{45.7}(13.5)&33.2(17.8)&\textbf{143.6}(32.2)&\textbf{134.9}(30.9)
        &\textbf{57.8}(13.4)&\uline{44.7}(30.2)&\textbf{41.9}(5.5)&\textbf{38.2}(4.9)\\
        \midrule
        \multirow{3}{*}{\textbf{128}}
        &JPS&\textbf{98.7}(2.0) &-&\textbf{155.0}(43.4) &147.4(10.0)
        &{47.7}(12.7) &-&\uline{431.0}(265.3)&401.3(258.8)
        &-111.9(27.6)&-&115.9(77.1)&95.3(13.0)\\
        &WA* & 65.2(6.2)& \uline{51.4}(13.0)& 167.5(13.1)&\uline{151.9}(12.0)
        &52.2(17.3)& 35.8(17.0)&432.3(257.1) & \uline{412.1}(262.5)
        &67.3(12.9)& \uline{56.5}(39.3)& \uline{106.3}(15.0)&\uline{100.9}(13.4)\\
        &Neural A*&67.2(6.6)&46.0(12.3) &173.0(13.4) &158.0(12.1)
        &\textbf{59.4}(11.7)&\textbf{42.7}(16.9) &442.3(261.7) &423.5(256.1)
        &\uline{71.1}(13.0) &55.4(41.7) &116.9(16.4)&111.6(14.7)\\
        &iA*(Ours) &\uline{67.6}(6.1) &\textbf{56.0}(12.2) &\uline{166.4}(11.3) &\textbf{151.4}(10.0)
        &\uline{54.0}(11.8) &\uline{39.5}(17.6) &\textbf{427.1}(263.0) &\textbf{406.3}(257.8)
        &\textbf{74.4}(11.5) &\textbf{60.0}(35.9) &\textbf{105.9}(14.9) &\textbf{100.6}(13.4)\\
        \midrule
        
        \multirow{3}{*}{\textbf{256}}
        &JPS&\textbf{99.4}(1.5) &-&\textbf{300.4}(83.2) & 277.9(25.8)
        &56.8(8.1) &-&744.6(444.2)& 703.4(426.7)
        &-204.1(526) &-&207.5(162.4)& 153.2(28.0)\\
        &WA* & 74.1(16.3) & 62.5(28.3)& \uline{340.5}(60.0)&\textbf{300.0}(47.5)
        & 63.9(16.7)& 31.2(36.0)& \uline{741.5}(435.6)&\uline{714.8}(423.4)
        &70.0(12.0)& \textbf{54.4}(55.6)& \uline{175.3}(36.2)&\uline{163.0}(31.7)\\
        &Neural A*&76.0(16.2) &\uline{64.7}(27.7) &345.6(61.6) &\uline{306.5}(46.9)
        &\uline{68.2}(9.7) &\uline{37.6}(21.7) &752.3(435.7)&727.0(423.1)
        &\textbf{74.0}(15.1) &\uline{53.3}(57.7) &194.2(41.9)&182.9(36.9)\\
        &iA*(Ours) &\uline{80.0}(13.4) &\textbf{69.7}(23.4) &342.9(62.0)&307.3(48.4)
        &\textbf{71.0}(12.2) &\textbf{39.5}(28.2) &\textbf{738.0}(435.6) & \textbf{711.8}(423.6)
        &\uline{71.2}(12.0)& 46.2(60.7)& \textbf{171.4}(35.2)&\textbf{159.5}(30.3)\\
        \bottomrule
        \end{tabular}
        }
        \vspace{-15pt}
    \end{table*}
    \subsection{Metrics}\label{sec:metrics}
    To evaluate the performance of the proposed method and facilitate comparison with the selected approaches,  we mainly use the following three metrics to compare their performance on search area reduction, runtime reduction, and the ability to balance search area and path length.
    \begin{itemize}
        \item \textbf{Reduction ratio of node explorations (\textit{Exp})}~\cite{yonetani2021nastar}: To measure the search efficiency of the examined method, we utilize the ratio of the reduced search area to the classical A* search as a metric, which is defined as 
        \begin{equation}
            \textit{Exp} = 100 \times \frac{A^* - A}{A^*},
            \label{eq:Exp}
        \end{equation}
        where $A$ represents the search area of the examined method and $A^*$ is the search area of classical A* search.
        
        \item \textbf{Reduction ratio of time (\textit{Rt})}: To measure the time efficiency of the examined method, we utilize the ratio of saved time to the operation time of standard A* search as a metric, which is defined as 
        \begin{equation}
            \textit{Rt} = 100 \times \frac{t^*-t}{t^*},
            \label{eq:Rt}
        \end{equation}
        where $t^*$ is the operating time of classical A* solution, and $t$ is the operating time of the examined method.
        
        \item \textbf{Sum of the search area and path length (\textit{AL})}: To measure the ability to balance the search area and path length, we use the combination of the search area and path length as a metric, which is defined as
        \begin{equation}
            \setlength{\abovedisplayskip}{2pt}
            \setlength{\belowdisplayskip}{2pt}
            \textit{AL} = \sqrt{\mathcal{L}_a(\mu)} + \mathcal{L}_l(\mu),
            \label{eq:AL}
        \end{equation}
        where $\mathcal{L}_a(\mu)$ and $\mathcal{L}_l(\mu)$ are the search area and path length, given path $\mu$. For a consistent unit, we use the square root of $\mathcal{L}_a(\mu)$ plus $\mathcal{L}_l(\mu)$ as the final metric.
        \item \textbf{Path Length} (\textit{PL}): To measure the path optimality of the examined method, we directly use the average of the absolute path length $\mathcal{L}_l(\mu)$ as a metric.
    \end{itemize}

    \subsection{Overall Performance}

    \textbf{Comparisons with Benchmarks}: To evaluate the path planning performance of iA*, we conduct both qualitative and quantitative experiments.
    \fref{fig:exp:benchmarks} visualizes the planning results of the selected methods. 
    From the results, the proposed iA* can greatly reduce the search area and still generate paths similar to optimal paths generated by classical A*.
    Furthermore, \tref{tab:exp:benchmarks} summarizes the quantitative experimental results across the MP, Maze, and Matterport3D datasets with three different map sizes $64\times64$, $128\times128$, and $256\times256$. 
    To ensure fairness, we conducted 10 randomized trials, totally including 80 instances, for each kind of map in all three datasets, sampling start and goal positions uniformly. 
    Final results are the average scores of these trials with the mentioned metrics. 
    Notably, the transformer-based FS + PPM and WA* + CF support only maps with a fixed shape, $64\times64$, because their network designs include fully connected layers that necessitate a predetermined input shape.  
    Adjustments and re-training procedures are required to meet varied input instance shapes. 
    Thus, comparisons at larger map sizes mainly focus on JPS, WA*, Neural A*, and iA*.
    Since there is no available batched JPS algorithm running on GPU, we exclude the $Rt$ metric in JPS results for fair comparisons.
    In the path length comparison, the results of the JPS algorithm are excluded, as JPS is designed to guarantee path optimality when using an admissible heuristic.

    From the quantitative experimental results on the MP dataset, it is evident that the JPS algorithm exhibits excellent performance in reducing the search area by approximately 98\% while maintaining path optimality. However, in the Matterport3D and Maze datasets, JPS performs poorly since jumps are constrained by frequent walls or turns, resulting in their poor performance on \textit{Exp} and \textit{AL}.
    The FS + PPM algorithm faces the similar limitation: it performs well on the MP dataset but deteriorates in the other two datasets.  
    Different from the JPS algorithm, FS + PPM is likely overfitting on the MP dataset  due to its powerful embedded neural network.
    The rest of the selected methods, WA*, Neural A*, WA* + CF, and iA* demonstrate consistent performance among the diverse maps with different sizes.  
    

    
    iA* has satisfactory performance among them. The proposed iA* gets the lowest \textit{AL} scores in 6 comparisons on 9 diverse environments with different map sizes.
    iA* saves on average reduction of 65.7\% in search area, 54.4\% in search time while still keeping the suboptimal paths with only a 3.8\% increase on path length.
    Compared with the baseline Neural A* , iA* reduces the average search area by 3.3\% and saves the runtime by 11.5\%. Meanwhile, it achieves a 6.3\% improvement in the trade-off between path length and search area, alongside a 6.1\% reduction in path length. 
    These results demonstrate that iA* can effectively minimize the search area while still identifying suboptimal paths.
    Notably, its consistent performance across different map sizes validates the method’s strong generalization without overfitting, which is critical for robotic applications.
    
    \begin{table}[t]
        \centering
        \caption{Quantitative Results in the Simulated Environments}
        \renewcommand{\arraystretch}{1.}
        \resizebox{\linewidth}{!}{
            \begin{tabular}{c|c|ccc}
                \toprule
                &  & A* & Neural A*& iA*\\
                \midrule
                \multirow{4}*{\rotatebox{90}{Indoor}} 
                & \textit{Exp} (Area)&0 (9826)&3.95\% (9438)&\textbf{43.28\%} (5573)\\ 
                & \textit{Rt} (Time)&0 (3.11 \textit{s})&3.67\% (2.99 \textit{s})&\textbf{52.04\%} (1.49 \textit{s})\\
                &\textit{AL}&473.17 &483.03	&\textbf{453.77}\\
                &\textit{PL}&374.04&385.88&\textbf{379.12}
                \\
                \midrule
                \multirow{4}*{\rotatebox{90}{Tunnel}}
                & \textit{Exp} (Area)&0 (45763)&13.37\% (39643)&\textbf{62.73\%} (17054)\\ 
                & \textit{Rt} (Time)&0 (14.31 \textit{s})&14.99\% (12.16 \textit{s})&\textbf{62.84\% }(5.31 \textit{s})\\
                & \textit{AL}&1014.94	&1021.34	&\textbf{939.99}\\
                &\textit{PL}&801.02&822.23&\textbf{809.40}
                \\
                \bottomrule
            \end{tabular}
        }
        \label{tab:exp:simu_exp}
        \vspace{-5pt}
    \end{table}
    
    \subsection{Autonomous Exploration Development Environments}
\begin{figure}[t]
        \centering
        \vspace{3pt}
        \includegraphics[width=0.95\linewidth]{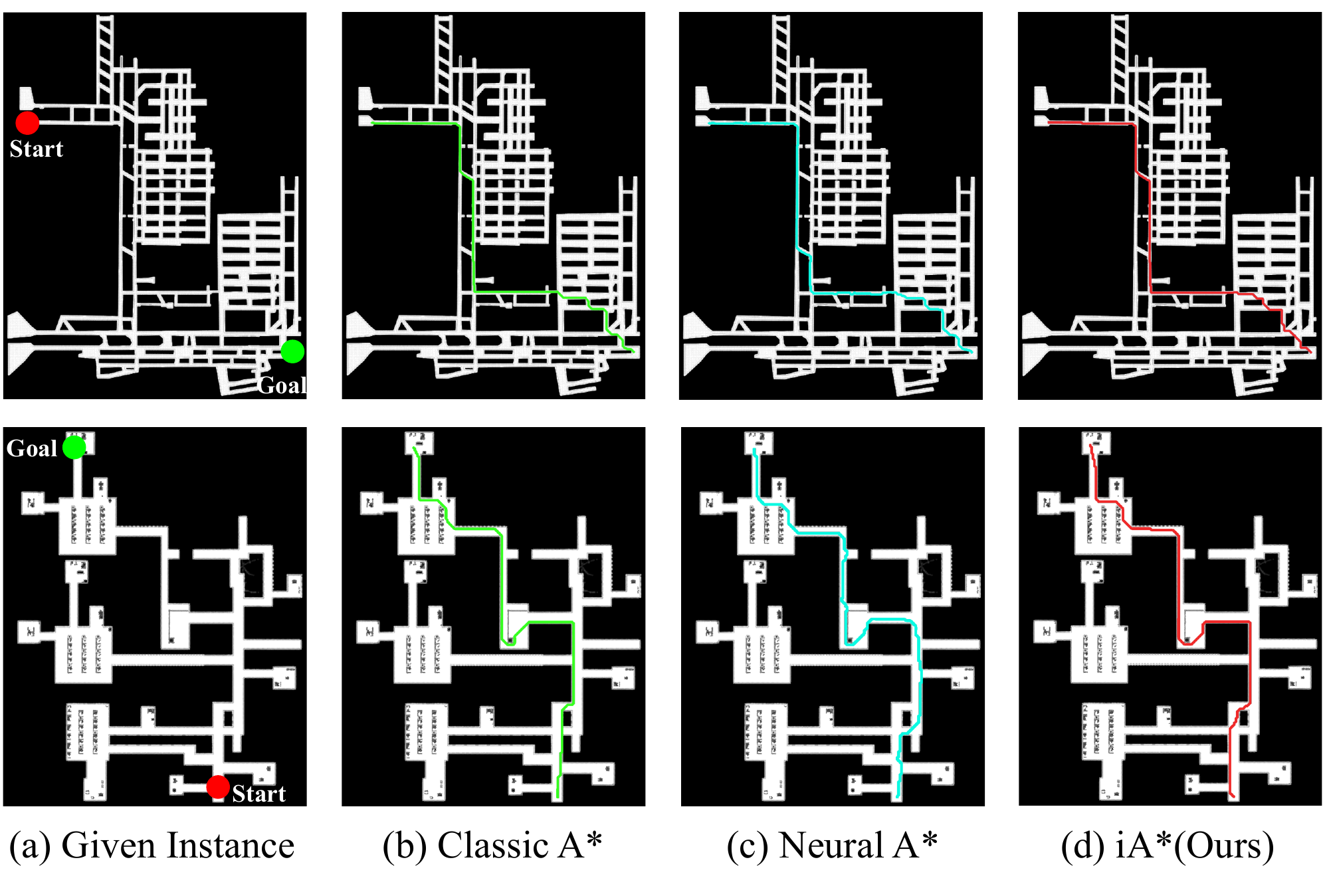}
        \caption{The trajectories of a mobile robot under given navigation tasks in the simulation environments, Tunnel (the first row) and Indoor (the second row). (a) represents the given instances, start (red circle), and goal (green circle). (b), (c) and (d) represent the trajectories of a mobile robot navigating from start to goal in the simulation environments with Classical A*, Neural A*, and iA*.}
        \label{fig:exp:simu_exp}
        \vspace{-10pt}
    \end{figure}

    Applying the proposed iA* in real world needs a complete environment map, a reliable localization algorithm, and a stable wheeled platform. Thus, to validate the practicality, we evaluate iA* and baseline methods in the widely used simulation Autonomous Exploration Development Environment ~\cite{cao2022Autonomous} that includes a mobile robot and several scenarios resembling real-world settings. 
    We use the indoor and the tunnel scenarios, shown as \fref{fig:exp:simu_exp} (a), as our test environments.
    Specifically, we convert their dense point cloud maps into two-dimensional binary maps, where each grid cell represents a $0.5 \meter \times 0.5 \meter$ area in the original map. 
    A grid cell will be regarded as an obstacle if its maximum $z$-value exceeds a given threshold; otherwise, it will be regarded as a free cell.
    Ultimately, the Tunnel and Indoor scenes have a size of $654 \times 505 $ and $259 \times 199$, respectively.
    Since these map sizes are different, Neural A* is the only applicable data-driven baseline method.
    As a result, we mainly compare the performance of A*, Neural A*, and iA*.
    It is worth noting that both the Neural A* and iA* use the MP dataset for training and never see these scenarios.

    In tests, the goal position for the mobile robot is manually given.
    We use the selected methods to generate paths and utilize the integrated path-following method within the maps to provide control commands for navigation.
    As shown in \fref{fig:exp:simu_exp}, both Neural A* and iA* can generate obstacle-free paths, similar to the optimal path generated by A*, and allow the robot to finish the task successfully. 
    As shown in \tref{tab:exp:simu_exp}, iA* significantly improves the planning efficiency compared to the baseline methods. 
    Specifically, in the Indoor scenario, iA* reduces 43.28\% search area and saves 52.04\% time; in the Tunnel scenario, it reduces 62.73\% search area and saves 62.84\% time. Besides, in these two scenarios, iA* gets lower \textit{AL} values, compared with Neural A*, presenting a better trade-off between search area and path length.

    As the map size increases by 6.4 times (from 259×199 to 654×505), traditional A* exhibits a 4.6-fold growth in search area and runtime, while iA* scales more efficiently with only a 3.3-fold increase. This highlights iA*'s superior adaptability to larger environments, maintaining faster computation.

    \subsection{Ablation Study}

    \begin{figure}[t]
        \centering
        \subfigure[A* + UNet\label{fig:AUNET}]{\centering\includegraphics[width=0.48\linewidth]{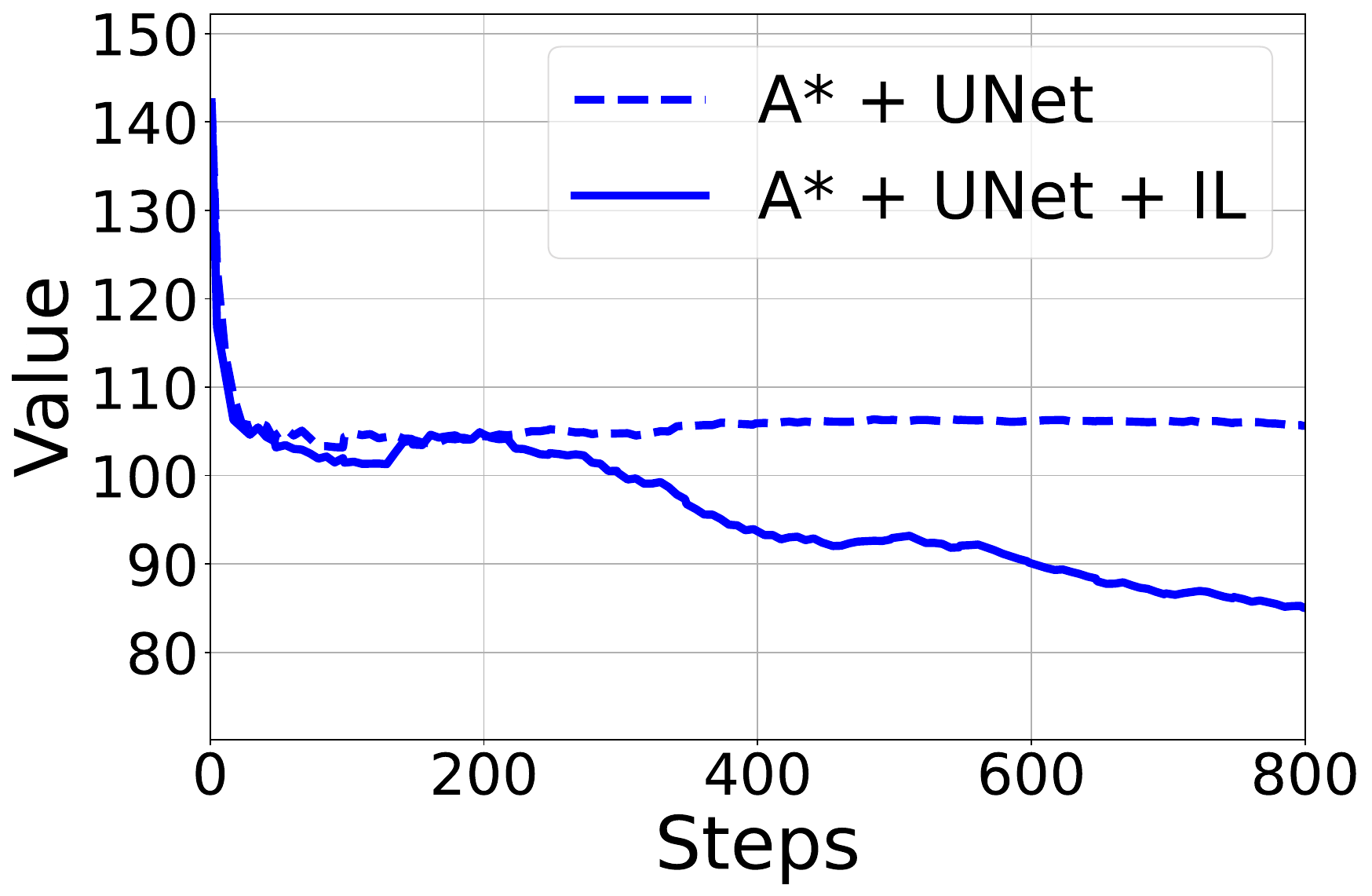}}
        \subfigure[A* + CNN\label{fig:ACNN}]{\centering\includegraphics[width=0.48\linewidth]{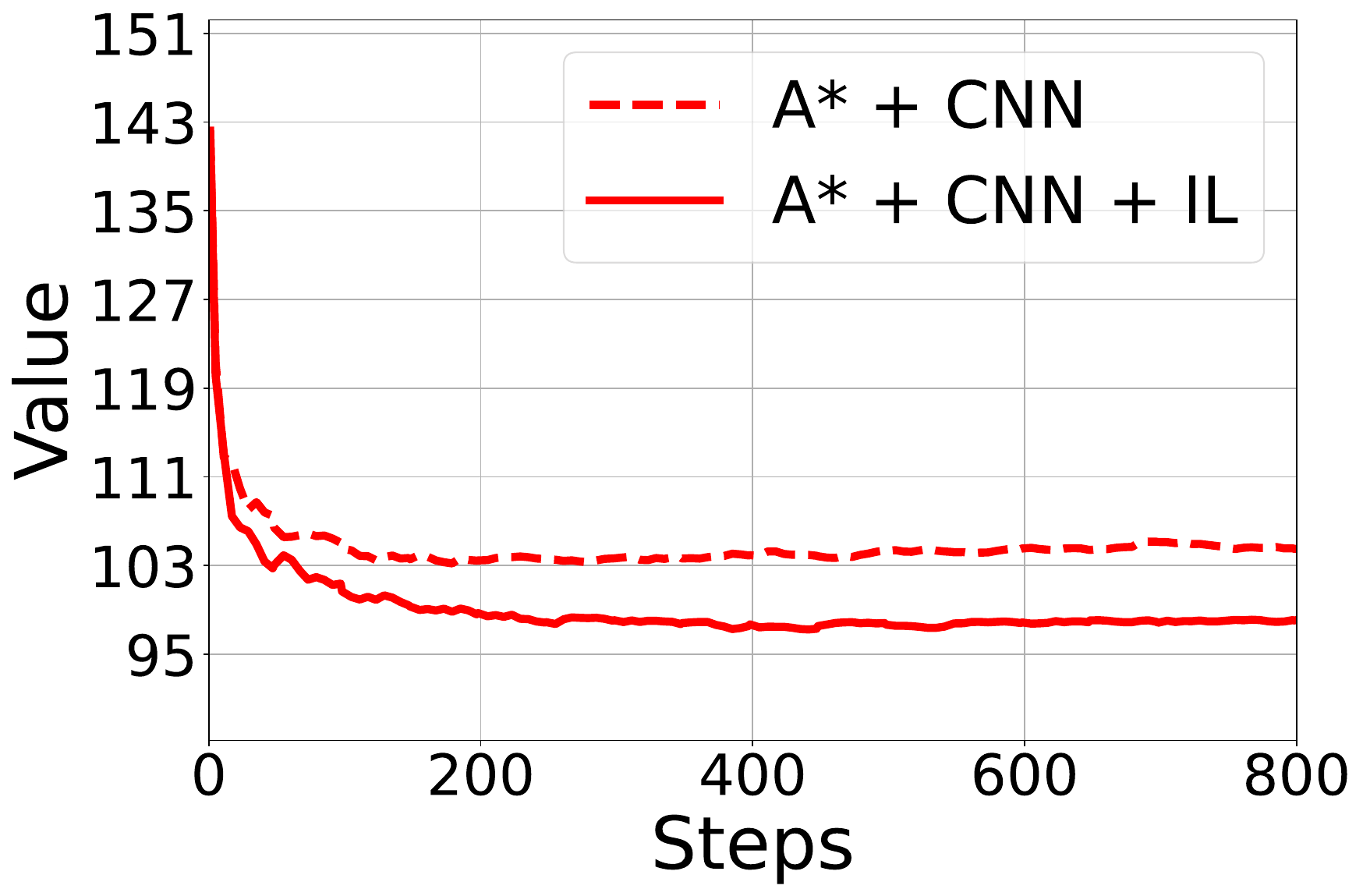}}
        \subfigure[WA* + UNet\label{fig:WAUNET}]{\centering\includegraphics[width=0.48\linewidth]{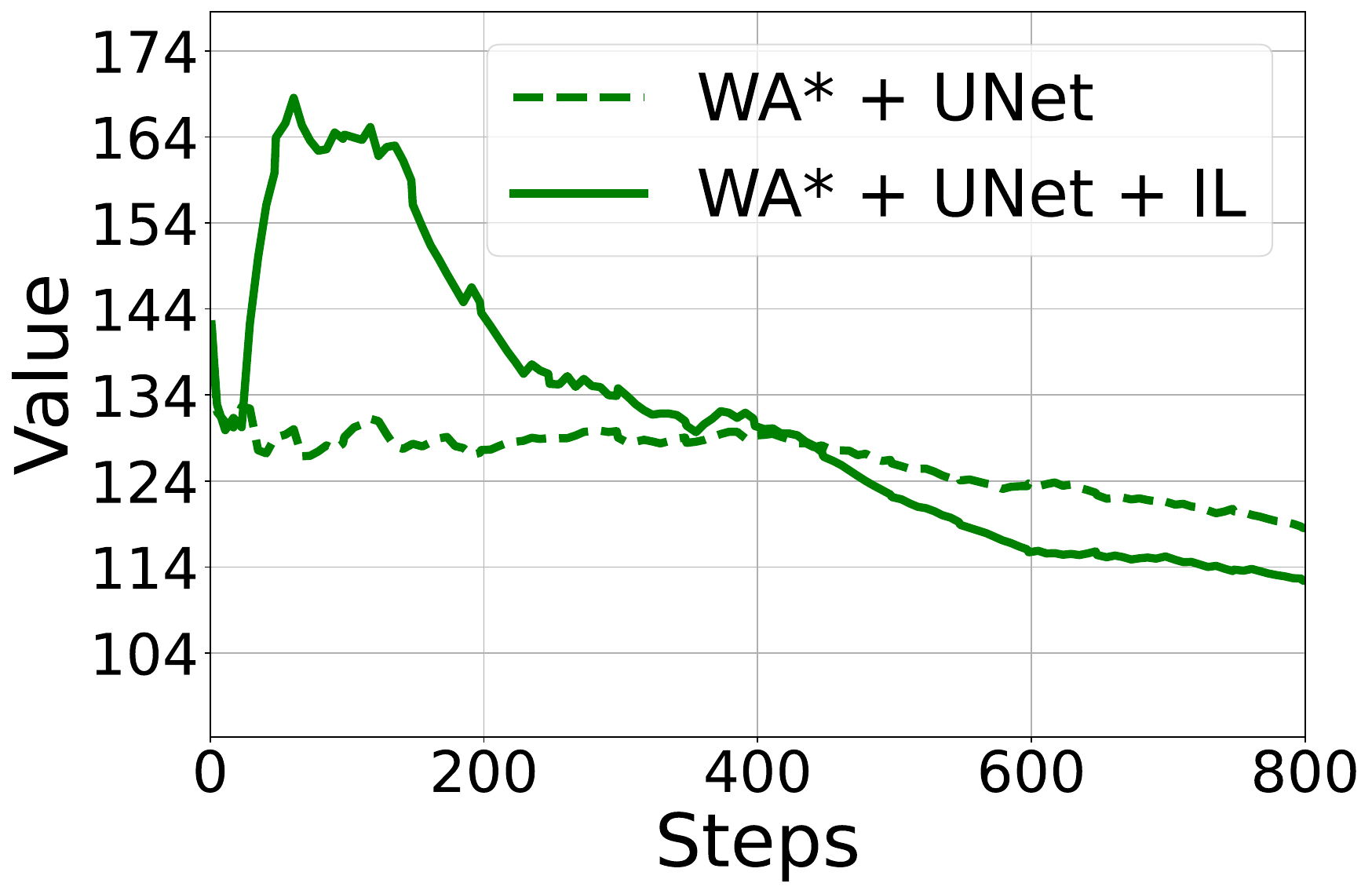}}
        \subfigure[WA* + CNN\label{fig:WACNN}]{\centering\includegraphics[width=0.48\linewidth]{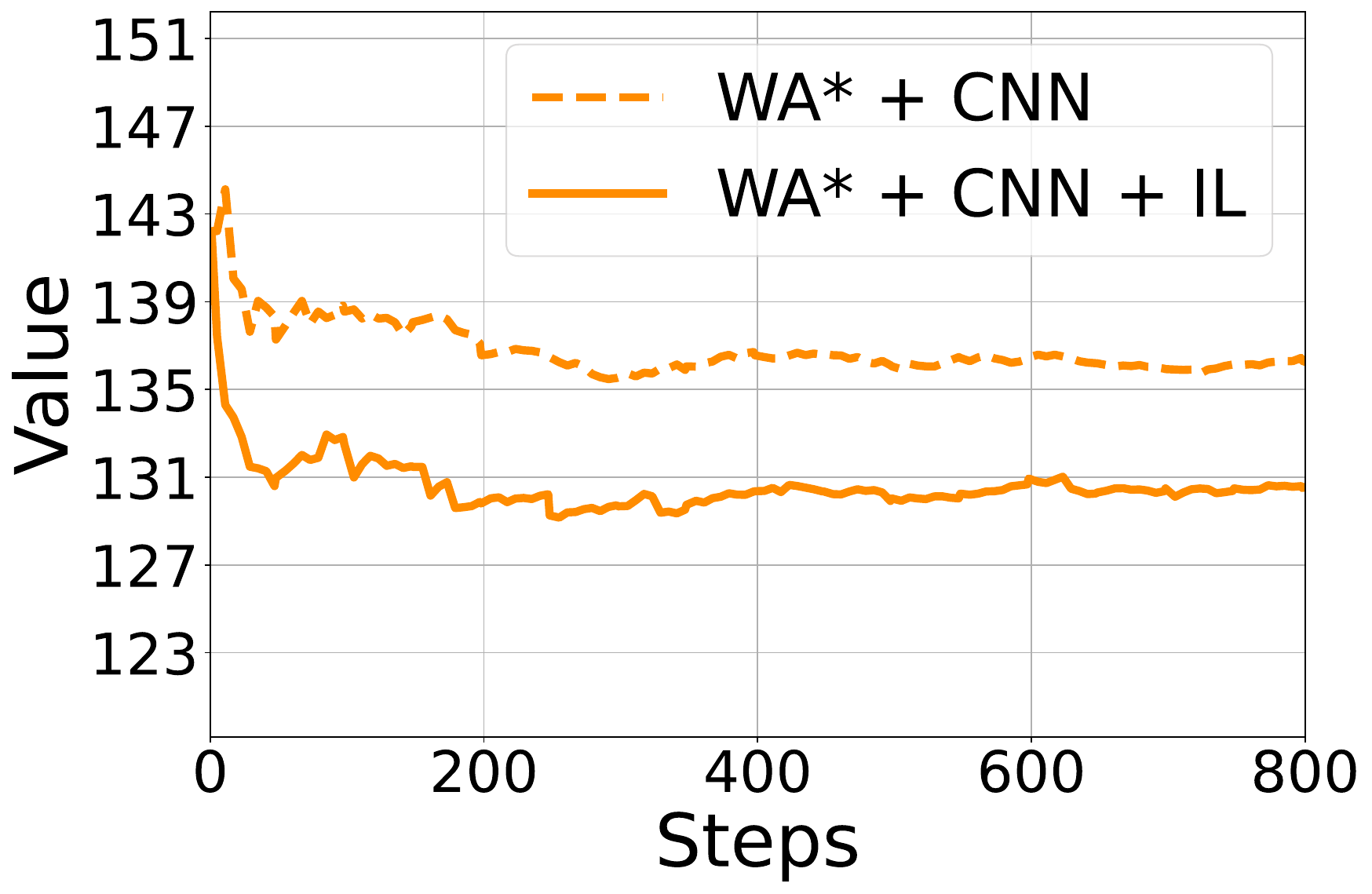}}
        \caption{Ablation study on imperative learning. (a) - (d) record the \textit{AL} curves of four different configurations in validation process with and without imperative learning. The four configurations are formed by pairing two search engines (A* and WA*) and two network architectures (CNN and UNet)
        The solid lines denote the methods with IL-based training, while dashed lines are those without IL.}
        \label{fig:ablation study}
        \vspace{-10pt}
    \end{figure}
    We next evaluate the effectiveness of the IL strategy under different configurations. 
    Four different configurations are considered by combining two search engines (A* and WA*) and two network architectures (CNN and UNet).
    Each configuration is trained both with and without the proposed IL training strategy. 
    For non-IL training methods, we follow the Neural A* framework, employing the ground truth optimal paths as supervision.
    All other training settings, including the learning rate and optimizer, are kept consistent to ensure a fair comparison.
    To assess the trade-off between path optimality and search efficiency, we track the variation of the \textit{AL} throughout the validation processes and show it in~\fref{fig:AUNET}-\ref{fig:WACNN}, where solid and dashed lines represent the methods with and without IL-based training, respectively. 
    As illustrated, the methods with IL consistently converge to significantly lower \textit{AL} values than their non-IL counterparts.
    On average, our IL-based methods achieve approximately a 9.79\% reduction in \textit{AL}, demonstrating the effectiveness of IL and balancing between path optimality and search efficiency.

    To further analyze the individual contributions of search engines (A* and WA*) and network architectures (CNN and UNet) under IL strategy, we isolate and compare the results (solid lines in \fref{fig:ablation study}) of the IL-trained configurations.
    The results show that models using the A* search engine (red and blue curves) outperform those employing WA* (green and yellow curves) in terms of lower \textit{AL} scores. 
    Moreover, the UNet-based models present superior performance compared to CNN-based models when paired with the same search engine.
    Consequently, based on these empirical observations, we adopt the configuration comprising the A* search algorithm and the UNet architecture for our final framework.

    \section{Conclusion and Future Work}

    We present iA*, an innovative search-based path planning framework designed to address critical challenges in existing methodologies. Leveraging imperative learning (IL), iA* integrates neural networks with a differentiable A* algorithm in a unified bilevel optimization framework, effectively eliminating the need for extensive labeled datasets through self-supervised training. This approach mitigates overfitting and significantly enhances generalization to previously unseen environments. Comprehensive experimental evaluations across diverse datasets and realistic simulation scenarios demonstrated iA*'s superior performance compared to both classical and supervised learning-based methods. The effectiveness of our IL-based training approach is further confirmed by ablation studies, underscoring its robustness and adaptability. 
    In future work, we aim to extend iA* to high-dimensional path planning problems, where the neural model may efficiently address the computational explosion of the search space. Furthermore, enhancing iA* with real-time replanning capabilities will broaden its applicability to dynamic environments, where fast adaptation and responsiveness are crucial for reliable performance.


{
    \small
    \balance
    \bibliographystyle{./bibliography/IEEEtran}
    \bibliography{reference}

@inproceedings{yonetani2021nastar,
  title={Path planning using neural a* search},
  author={Yonetani, Ryo and Taniai, Tatsunori and Barekatain, Mohammadamin and Nishimura, Mai and Kanezaki, Asako},
  booktitle={International conference on machine learning},
  pages={12029--12039},
  year={2021},
  organization={PMLR}
}

@article{hart1968formal,
  title={A formal basis for the heuristic determination of minimum cost paths},
  author={Hart, Peter E and Nilsson, Nils J and Raphael, Bertram},
  journal={IEEE transactions on Systems Science and Cybernetics},
  volume={4},
  number={2},
  pages={100--107},
  year={1968},
  publisher={IEEE}
}

@inproceedings{kirilenko2023transpath,
  title={TransPath: learning heuristics for grid-based pathfinding via transformers},
  author={Kirilenko, Daniil and Andreychuk, Anton and Panov, Aleksandr and Yakovlev, Konstantin},
  booktitle={Proceedings of the AAAI Conference on Artificial Intelligence},
  volume={37},
  number={10},
  pages={12436--12443},
  year={2023}
}

@article{choudhury2018data,
  title={Data-driven planning via imitation learning},
  author={Choudhury, Sanjiban and Bhardwaj, Mohak and Arora, Sankalp and Kapoor, Ashish and Ranade, Gireeja and Scherer, Sebastian and Dey, Debadeepta},
  journal={The International Journal of Robotics Research},
  volume={37},
  number={13-14},
  pages={1632--1672},
  year={2018},
  publisher={SAGE Publications Sage UK: London, England}
}

@article{dijkstra1959note,
  title={A note on two problems in connexion with graphs},
  author={Dijkstra, EW},
  journal={Numerische Mathematik},
  volume={1},
  number={1},
  pages={269--271},
  year={1959},
  publisher={Springer-Verlag Berlin, Heidelberg}
}

@article{pohl1970heuristic,
  title={Heuristic search viewed as path finding in a graph},
  author={Pohl, Ira},
  journal={Artificial intelligence},
  volume={1},
  number={3-4},
  pages={193--204},
  year={1970},
  publisher={Elsevier}
}

@inproceedings{bhardwaj2017learning,
  title={Learning heuristic search via imitation},
  author={Bhardwaj, Mohak and Choudhury, Sanjiban and Scherer, Sebastian},
  booktitle={Conference on Robot Learning},
  pages={271--280},
  year={2017},
  organization={PMLR}
}

@article{maze-dataset,
  title={A Configurable Library for Generating and Manipulating Maze Datasets},
  author={Ivanitskiy, Michael Igorevich and Shah, Rusheb and Spies, Alex F and R{\"a}uker, Tilman and Valentine, Dan and Rager, Can and Quirke, Lucia and Mathwin, Chris and Corlouer, Guillaume and Behn, Cecilia Diniz and others},
  journal={arXiv preprint arXiv:2309.10498},
  year={2023}
}

@inproceedings{chang2017matterport3d,
  title={Matterport3D: Learning from RGB-D Data in Indoor Environments},
  author={Chang, Angel and Dai, Angela and Funkhouser, Thomas and Halber, Maciej and Niebner, Matthias and Savva, Manolis and Song, Shuran and Zeng, Andy and Zhang, Yinda},
  booktitle={2017 International Conference on 3D Vision (3DV)},
  pages={667--676},
  year={2017},
  organization={IEEE Computer Society}
}

@article{lavalle2001rapidly,
  title={Rapidly-exploring random trees: Progress and prospects},
  author={LaValle, Steven M and Kuffner, James J and Donald, BR and others},
  journal={Algorithmic and computational robotics: new directions},
  volume={5},
  pages={293--308},
  year={2001},
  publisher={Wellesley}
}

@article{strub2022adaptively,
  title={Adaptively informed trees (AIT*) and effort informed trees (EIT*): Asymmetric bidirectional sampling-based path planning},
  author={Strub, Marlin P and Gammell, Jonathan D},
  journal={The International Journal of Robotics Research},
  volume={41},
  number={4},
  pages={390--417},
  year={2022},
  publisher={SAGE Publications Sage UK: London, England}
}

@article{gammell2020batch,
  title={Batch Informed Trees (BIT*): Informed asymptotically optimal anytime search},
  author={Gammell, Jonathan D and Barfoot, Timothy D and Srinivasa, Siddhartha S},
  journal={The International Journal of Robotics Research},
  volume={39},
  number={5},
  pages={543--567},
  year={2020},
  publisher={SAGE Publications Sage UK: London, England}
}

@inproceedings{yang2023iplanner,
  author = {Yang, Fan and Wang, Chen and Cadena, Cesar and Hutter, Marco},
  title = {{iPlanner}: Imperative Path Planning},
  booktitle = {Robotics: Science and Systems (RSS)},
  year = {2023},
}

@inproceedings{zhan2024imatching,
  title = {{iMatching}: Imperative Correspondence Learning},
  author = {Zhan, Zitong and Gao, Dasong and Lin, Yun-Jou and Xia, Youjie and Wang, Chen},
  booktitle = {European Conference on Computer Vision (ECCV)},
  year = {2024}
}

@article{fu2023islam,
  title = {{iSLAM}: Imperative {SLAM}},
  author = {Fu, Taimeng and Su, Shaoshu and Lu, Yiren and Wang, Chen},
  journal = {IEEE Robotics and Automation Letters (RA-L)},
  year = {2024},
  code = {https://github.com/sair-lab/iSLAM/},
  video = {https://www.youtube.com/watch?v=rtCvx0XCRno},
  addendum = {SAIR Lab Recommended}
}

@book{horn2012matrix,
  title={Matrix analysis},
  author={Horn, Roger A and Johnson, Charles R},
  year={2012},
  publisher={Cambridge university press}
}

@book{west2001introduction,
  title={Introduction to graph theory},
  author={West, Douglas Brent and others},
  volume={2},
  year={2001},
  publisher={Prentice hall Upper Saddle River}
}

@article{doran1966experiments,
  title={Experiments with the graph traverser program},
  author={Doran, James E and Michie, Donald},
  journal={Proceedings of the Royal Society of London. Series A. Mathematical and Physical Sciences},
  volume={294},
  number={1437},
  pages={235--259},
  year={1966},
  publisher={The Royal Society London}
}

@inproceedings{gammell2014informed,
  title={Informed RRT: Optimal sampling-based path planning focused via direct sampling of an admissible ellipsoidal heuristic},
  author={Gammell, Jonathan D and Srinivasa, Siddhartha S and Barfoot, Timothy D},
  booktitle={2014 IEEE/RSJ international conference on intelligent robots and systems},
  pages={2997--3004},
  year={2014},
  organization={IEEE}
}

@article{liu2023prm,
  title={PRM-D* Method for Mobile Robot Path Planning},
  author={Liu, Chunyang and Xie, Saibao and Sui, Xin and Huang, Yan and Ma, Xiqiang and Guo, Nan and Yang, Fang},
  journal={Sensors},
  volume={23},
  number={7},
  pages={3512},
  year={2023},
  publisher={MDPI}
}

@inproceedings{bohlin2000path,
  title={Path planning using lazy PRM},
  author={Bohlin, Robert and Kavraki, Lydia E},
  booktitle={Proceedings 2000 ICRA. Millennium conference. IEEE international conference on robotics and automation. Symposia proceedings (Cat. No. 00CH37065)},
  volume={1},
  pages={521--528},
  year={2000},
  organization={IEEE}
}

@article{ravankar2020hpprm,
  title={HPPRM: hybrid potential based probabilistic roadmap algorithm for improved dynamic path planning of mobile robots},
  author={Ravankar, Ankit A and Ravankar, Abhijeet and Emaru, Takanori and Kobayashi, Yukinori},
  journal={IEEE Access},
  volume={8},
  pages={221743--221766},
  year={2020},
  publisher={IEEE}
}

@article{wang2020neural,
  title={Neural RRT*: Learning-based optimal path planning},
  author={Wang, Jiankun and Chi, Wenzheng and Li, Chenming and Wang, Chaoqun and Meng, Max Q-H},
  journal={IEEE Transactions on Automation Science and Engineering},
  volume={17},
  number={4},
  pages={1748--1758},
  year={2020},
  publisher={IEEE}
}

@inproceedings{takahashi2019learning,
  title={Learning heuristic functions for mobile robot path planning using deep neural networks},
  author={Takahashi, Takeshi and Sun, He and Tian, Dong and Wang, Yebin},
  booktitle={Proceedings of the International Conference on Automated Planning and Scheduling},
  volume={29},
  pages={764--772},
  year={2019}
}

@article{vaswani2017attention,
  title={Attention is all you need},
  author={Vaswani, Ashish and Shazeer, Noam and Parmar, Niki and Uszkoreit, Jakob and Jones, Llion and Gomez, Aidan N and Kaiser, {\L}ukasz and Polosukhin, Illia},
  journal={Advances in neural information processing systems},
  volume={30},
  year={2017}
}

@inproceedings{ronneberger2015u,
  title={U-net: Convolutional networks for biomedical image segmentation},
  author={Ronneberger, Olaf and Fischer, Philipp and Brox, Thomas},
  booktitle={Medical Image Computing and Computer-Assisted Intervention--MICCAI 2015: 18th International Conference, Munich, Germany, October 5-9, 2015, Proceedings, Part III 18},
  pages={234--241},
  year={2015},
  organization={Springer}
}

@article{qureshi2020motion,
  title={Motion planning networks: Bridging the gap between learning-based and classical motion planners},
  author={Qureshi, Ahmed Hussain and Miao, Yinglong and Simeonov, Anthony and Yip, Michael C},
  journal={IEEE Transactions on Robotics},
  volume={37},
  number={1},
  pages={48--66},
  year={2020},
  publisher={IEEE}
}

@article{kaelbling1996reinforcement,
  title={Reinforcement learning: A survey},
  author={Kaelbling, Leslie Pack and Littman, Michael L and Moore, Andrew W},
  journal={Journal of artificial intelligence research},
  volume={4},
  pages={237--285},
  year={1996}
}

@article{panov2018grid,
  title={Grid path planning with deep reinforcement learning: Preliminary results},
  author={Panov, Aleksandr I and Yakovlev, Konstantin S and Suvorov, Roman},
  journal={Procedia computer science},
  volume={123},
  pages={347--353},
  year={2018},
  publisher={Elsevier}
}

@inproceedings{gao2019global,
  title={A global path planning algorithm for robots using reinforcement learning},
  author={Gao, Penggang and Liu, Zihan and Wu, Zongkai and Wang, Donglin},
  booktitle={2019 IEEE International Conference on Robotics and Biomimetics (ROBIO)},
  pages={1693--1698},
  year={2019},
  organization={IEEE}
}

@INPROCEEDINGS{cao2022Autonomous,
  author={Cao, Chao and Zhu, Hongbiao and Yang, Fan and Xia, Yukun and Choset, Howie and Oh, Jean and Zhang, Ji},
  booktitle={2022 International Conference on Robotics and Automation (ICRA)}, 
  title={Autonomous Exploration Development Environment and the Planning Algorithms}, 
  year={2022},
  volume={},
  number={},
  pages={8921-8928},
  doi={10.1109/ICRA46639.2022.9812330}}

@INPROCEEDINGS{dstar,
  author={Stentz, A.},
  booktitle={Proceedings of the 1994 IEEE International Conference on Robotics and Automation}, 
  title={Optimal and efficient path planning for partially-known environments}, 
  year={1994},
  volume={},
  number={},
  pages={3310-3317 vol.4},
  doi={10.1109/ROBOT.1994.351061}}

@article{review,
  title={A review: On path planning strategies for navigation of mobile robot},
  author={Patle, BK and Pandey, Anish and Parhi, DRK and Jagadeesh, AJDT and others},
  journal={Defence Technology},
  volume={15},
  number={4},
  pages={582--606},
  year={2019},
  publisher={Elsevier}
}

@article{Quan2020Survey,
  title={Survey of UAV motion planning},
  author={Quan, Lun and Han, Luxin and Zhou, Boyu and Shen, Shaojie and Gao, Fei},
  journal={IET Cyber-systems and Robotics},
  volume={2},
  number={1},
  pages={14--21},
  year={2020},
  publisher={Wiley Online Library}
}

@book{russell2016artificial,
  title={Artificial intelligence: a modern approach},
  author={Russell, Stuart J and Norvig, Peter},
  year={2016},
  publisher={Pearson}
}

@article{gonzalez2015review,
  title={A review of motion planning techniques for automated vehicles},
  author={Gonz{\'a}lez, David and P{\'e}rez, Joshu{\'e} and Milan{\'e}s, Vicente and Nashashibi, Fawzi},
  journal={IEEE Transactions on intelligent transportation systems},
  volume={17},
  number={4},
  pages={1135--1145},
  year={2015},
  publisher={IEEE}
}

@inproceedings{stentz1994optimal,
  title={Optimal and efficient path planning for partially-known environments},
  author={Stentz, Anthony},
  booktitle={Proceedings of the 1994 IEEE international conference on robotics and automation},
  pages={3310--3317},
  year={1994},
  organization={IEEE}
}

@article{wang2025imperative,
  title={Imperative learning: A self-supervised neuro-symbolic learning framework for robot autonomy},
  author={Wang, Chen and Ji, Kaiyi and Geng, Junyi and Ren, Zhongqiang and Fu, Taimeng and Yang, Fan and Guo, Yifan and He, Haonan and Chen, Xiangyu and Zhan, Zitong and others},
  journal={The International Journal of Robotics Research (IJRR)},
  year={2025},
  publisher={SAGE Publications Sage UK: London, England}
}

@inproceedings{li2025ikap,
  title = {{iKap}: Kinematics-aware Planning with Imperative Learning},
  author = {Li, Qihang and Chen, Zhuoqun and Zheng, Haoze and He, Haonan and Su, Shaoshu and Geng, Junyi and Wang, Chen},
  booktitle = {IEEE International Conference on Robotics and Automation (ICRA)},
  year = {2025}
}

@inproceedings{harabor2011jps,
  title={Online graph pruning for pathfinding on grid maps},
  author={Harabor, Daniel and Grastien, Alban},
  booktitle={Proceedings of the AAAI conference on artificial intelligence},
  volume={25},
  number={1},
  pages={1114--1119},
  year={2011}
}
}	
\end{document}